\documentclass[10pt,conference]{IEEEtran}
\usepackage{graphicx}
\usepackage[utf8]{inputenc}
\usepackage{amssymb,amsmath,array}

\usepackage[hidelinks]{hyperref}
\usepackage{orcidlink}

\begin{document}
\title{Evolutionary Warm-Starts for Reinforcement Learning in Industrial Continuous Control}


\author{
    \IEEEauthorblockN{Tom Maus \orcidlink{0009-0005-3496-5964}}
    \IEEEauthorblockA{\textit{Inst. for Neural Computation} \\
    \textit{Ruhr University Bochum}\\
    Bochum, Germany \\
    tom.maus@ini.rub.de}
    \and
    \IEEEauthorblockN{Stephan Frank \orcidlink{0009-0008-0112-6023}}
    \IEEEauthorblockA{\textit{Inst. for Neural Computation} \\
    \textit{Ruhr University Bochum}\\
    Bochum, Germany \\
    stephan.frank@ini.rub.de}
    \and
    \IEEEauthorblockN{Tobias Glasmachers \orcidlink{0000-0003-1886-1696}}
    \IEEEauthorblockA{\textit{Inst. for Neural Computation} \\
    \textit{Ruhr University Bochum}\\
    Bochum, Germany \\
    tobias.glasmachers@ini.rub.de}
}


\maketitle

\begin{abstract}
Reinforcement learning (RL) is still rarely applied in industrial control, partly due to the difficulty of training reliable agents for real-world conditions. This work investigates how evolution strategies can support RL in such settings by introducing a continuous-control adaptation of an industrial sorting benchmark. The CMA-ES algorithm is used to generate high-quality demonstrations that warm-start RL agents. Results show that CMA-ES-guided initialization significantly improves stability and performance. Furthermore, the demonstration trajectories generated with the CMA-ES provide a strong oracle reference performance level, which is of interest in its own right. The study delivers a focused proof of concept for hybrid evolutionary-RL approaches and a basis for future, more complex industrial applications.
\end{abstract}

\section{Introduction}
Reinforcement learning is increasingly explored for industrial process control, where systems exhibit stochastic dynamics, nonlinear interactions, and strict operational constraints \cite{cronrath_enhancing_2019, pendyala_containergym_2023, maus_sortingenv_2025}. While simulations and digital twins enable safe and flexible training, many RL studies still rely on abstract benchmarks with limited resemblance to real production systems, which restricts transferability to practice \cite{cronrath_enhancing_2019}. Recent industrially grounded environments, such as the Industrial Benchmark, ContainerGym, and SortingEnv, have begun to address this gap by incorporating quality targets, stochastic disturbances, and sequential process dependencies \cite{maus_sortingenv_2025, hein_benchmark_2017, pendyala_containergym_2023}.

Parallel to these developments, evolution strategies (ES) have gained traction as a complementary tool in RL for continuous black-box control and policy search, offering robustness in noisy domains \cite{hansen_completely_2001, hansen_method_2009, salimans_evolution_2017}. In the context of industrial sorting, GAs have been used to generate optimized trajectories for a discrete sensor-switching task, providing demonstration data that significantly improved Proximal Policy Optimization (PPO) training via behavioral cloning (BC) \cite{maus_leveraging_2025}.

However, real sorting facilities often rely on continuous controls (e.g., conveyor speeds), while the existing evolutionary demonstration-generation targets discrete actions. Direct GA-based trajectory search in continuous spaces is computationally prohibitive due to high dimensionality and long horizons, creating a gap between benchmarks and industrial reality \cite{maus_leveraging_2025}.

This work addresses this gap by adapting an industrial sorting benchmark into a continuous-control formulation focused on sequential input regulation. The environment retains the industrial motivation and underlying process logic of the original benchmark \cite{maus_leveraging_2025} but is substantially restructured to support a continuous control signal, including a redesigned reward function that jointly reflects quality and throughput objectives. Demonstration generation is cast as a continuous black-box optimization problem, solved using the Covariance Matrix Adaptation Evolution Strategy (CMA-ES) \cite{hansen_completely_2001, hansen_method_2009, salimans_evolution_2017}. By optimizing bounded control schedules under stochastic dynamics and industrial constraints, CMA-ES produces high-performing, seed-specific trajectories that serve as expert demonstrations to warm-start PPO training.

The contributions of this study are as follows: (i) a continuous-control reformulation of an industrially inspired sorting task, (ii) a systematic use of CMA-ES as an offline oracle for demonstration generation under realistic constraints, and (iii) an empirical evaluation showing how evolutionary warm-starting improves PPO performance and sample efficiency in comparison to self-exploring PPO agents. 
The results provide evidence that evolution strategies can substantially support RL training in industrially relevant continuous-control tasks and offer a foundation for further applied research on hybrid evolutionary RL methods in more complex industrial environments. The code used for the experiments in this study is publicly available. \footnote{\url{https://github.com/Storm-131/EvoRL}}

\section{The Environment and Problem Formulation}

\begin{figure*}[t!]
\centering
\includegraphics[width=1.0\textwidth]{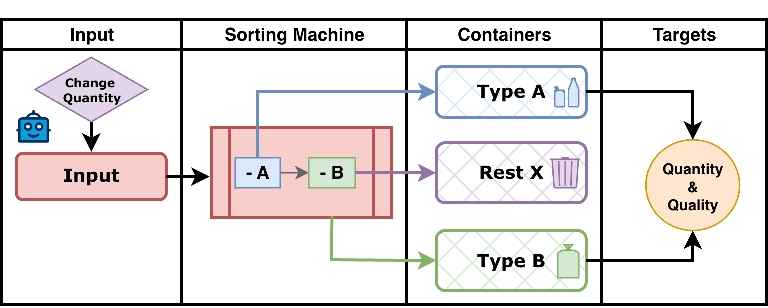}
\caption{Simplified schematic of the continuous sorting process, based on \cite{maus_sortingenv_2025, maus_leveraging_2025}. The agent regulates the input stream into the sorting machine for separation of materials. Output quantities and purities define the performance targets.
}\label{Fig:Schematic}
\end{figure*}

The environment used in this study is a continuous-control adaptation of an existing industrial sorting benchmark \cite{maus_leveraging_2025}. Its overarching objective is to regulate the input quantity of mixed recyclable material such that the system achieves an optimal balance between throughput and purity. This is essential because sorting efficiency deteriorates under high load, and the input composition varies over time. The underlying sorting process follows a fixed two-stage structure (see Fig. \ref{Fig:Schematic}). In Stage A, material of type A is separated with load-dependent accuracy. The residual flow enters Stage B, where type B is extracted with its own accuracy profile. Any remaining material is collected as a residual stream, representing misclassified or unsorted fractions. Misclassifications at either stage accumulate in the output containers and directly affect the purity of the respective material streams.

The original SortingEnv modeled a full multi-stage sorting and redistribution pipeline based on a binary sensor-selection mechanism \cite{maus_leveraging_2025}. In contrast, the present formulation restructures the system to its central regulatory sub-process and exposes a one-dimensional continuous control variable. The agent's action determines the normalized input quantity for each batch, thereby directly determining throughput and indirectly influencing purity through load-dependent sorting accuracies. This creates the fundamental trade-off between processing more material and maintaining sufficient quality. Implemented in Gymnasium \cite{towers_gymnasium_2023}, the environment models a simplified material-handling system in which the agent must maintain a target purity under fluctuating input conditions by continuously adjusting a bounded control parameter.

Each episode begins with a stochastically generated inflow produced by the input generator, which samples batch sizes and material compositions according to predefined sampling modes. These modes introduce variability and simple trend profiles. Stochasticity in the system originates from this input generation process. Each timestep consists of (i) sampling a new input batch, (ii) updating internal states, (iii) applying the agent's continuous action, (iv) evaluating purity and throughput, and (v) computing the reward before returning the next observation.

The action space consists of a single continuous variable mapped into a physically valid operating range. The observation space is a 7-dimensional continuous state vector composed of the last five measured input mixture ratios (capturing short-term fluctuations in the incoming A/B composition) and the current purities of output streams A and B.

The reward function incentivizes stable high-purity operation while accounting for throughput. Purity results from nonlinear load-dependent sorting accuracies, where misclassifications accumulate in the material containers over time. This accumulation creates a gradual evolution of purity across timesteps and introduces temporal dependencies that reflect the long-horizon nature of industrial sorting processes. Quality thresholds define the acceptable purity range for each material stream. Falling below a threshold incurs a strong penalty, while exceeding it provides a small positive bonus that decays toward zero as purity approaches unity. Throughput provides an additional quantity-dependent contribution. The overall reward at time $t$ is given in a compact form by:

\begin{equation}
\label{eq:reward_main}
r_t = 0.25 \left( 2 \frac{q_t}{q_{\max}} - 1 \right) + \sum_{i \in \{A,B\}} f(p_{i,t}, \theta_i)
\end{equation}
where the purity-dependent reward $f$ is given by:
\begin{equation}
f(p_{i,t}, \theta_i) = 
\begin{cases}
-10.0 + 10.25 \frac{p_{i,t}}{\theta_i}, & p_{i,t} < \theta_i \\
0.25 \left( 1 - \frac{p_{i,t}-\theta_i}{1-\theta_i} \right), & p_{i,t} \geq \theta_i
\end{cases}
\end{equation}

\noindent where $q_t$ denotes the current quantity, $q_{max}$ the nominal maximum capacity, $p_{i,t}$ the purity of stream $i$, and $\theta_i$ its purity threshold. This formulation yields a smooth, informative reward landscape with strong penalties for quality violations, making it suitable for both reinforcement learning and evolutionary optimization.

\begin{figure*}[t!]
\centering
\includegraphics[width=1.0\textwidth]{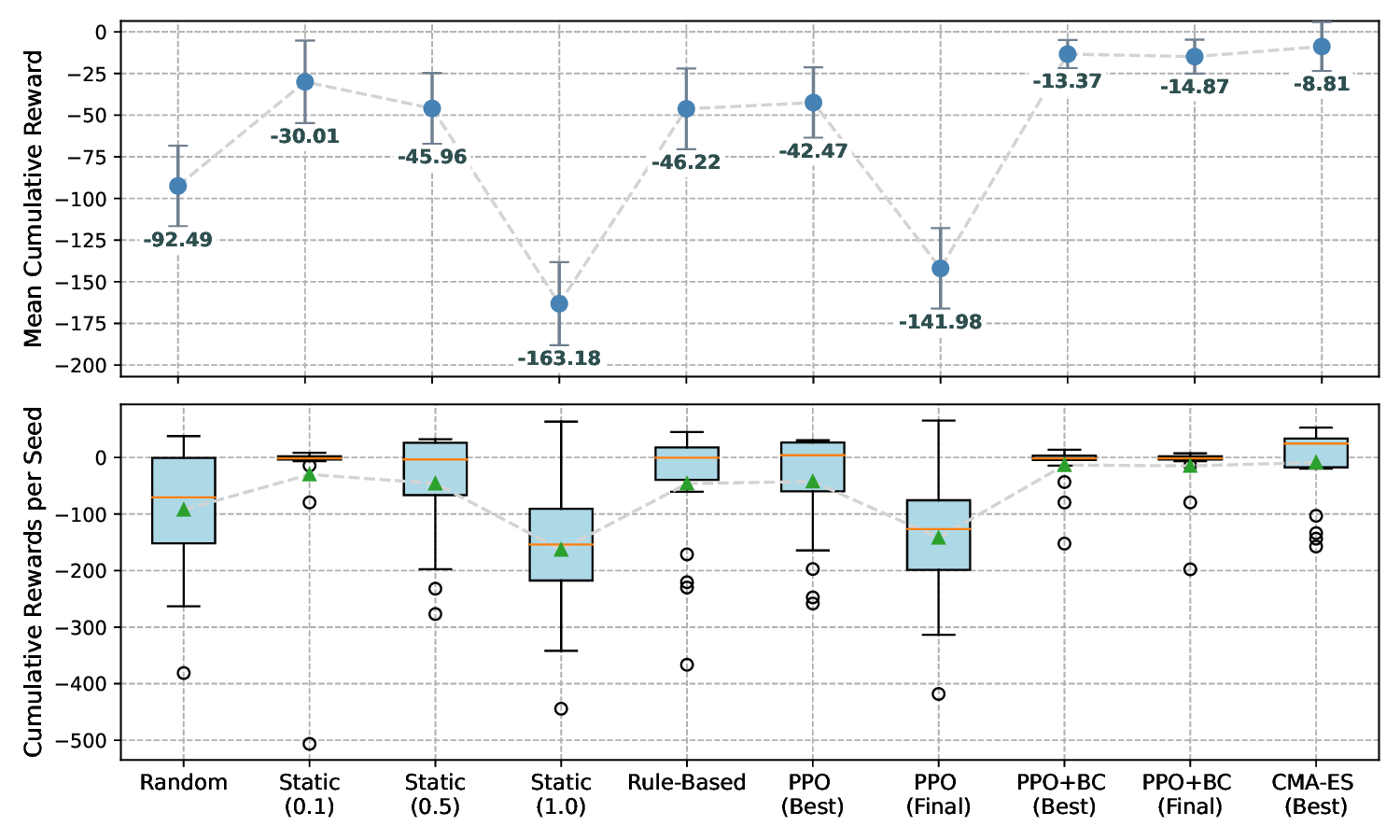}
\caption{Performance comparison across 20 test seeds with mean cumulative reward, standard deviation (top) and reward distributions (bottom).}\label{Fig:Results}
\end{figure*}

\section{CMA-ES-Based Demonstration Generation}
To generate high-quality demonstrations, we treat the control problem as a black-box trajectory optimization task solved by the CMA-ES \cite{hansen_completely_2001}. The optimization variable is the full episode-long sequence of continuous actions. Fitness values are yielded by the final cumulative reward for the action sequence. Each candidate is evaluated on a fixed environment with frozen random number generator seed, transforming the stochastic dynamics into a deterministic objective and effectively providing the optimizer with oracle-like access to the future input sequence. This enables the CMA-ES to discover seed-specific oracle trajectories which rely on complete foresight and therefore serve as an empirical upper bound. Aggregating optimized trajectories from multiple training seeds, we obtain a near-optimal demonstration set and use it to warm-start PPO via supervised Behavioral Cloning (PPO+BC) before standard RL fine-tuning.

\section{Experiments and Results}

The experimental evaluation aims to quantify the performance gain of CMA-ES-based warm-starting over standard RL and classical heuristics. In this section, we describe the experimental setup along with the hyperparameters used in detail.

To ensure reproducibility and address potential bias, all experiments were conducted using a fixed set of hyperparameters (see Table~\ref{tab:hyperparameters}). The RL agents were implemented using Stable-Baselines3 \cite{raffin_stable-baselines3_2021} with a lightweight MLP architecture ($[32, 32]$) to prevent overfitting on the 7-dimensional observation space. 

\begin{table}[h]
\caption{Hyperparameters for PPO, CMA-ES, and BC}
\label{tab:hyperparameters}
\centering
\begin{tabular}{ll|ll}
\hline
\textbf{PPO Parameter} & \textbf{Value} & \textbf{CMA-ES / BC} & \textbf{Value} \\ \hline
Learning Rate & $3 \cdot 10^{-4}$ & Population Size ($\lambda$) & $16$ \\
Policy Arch. (MLP) & $[32, 32]$ & Max. Generations & $30$ \\
Value Arch. (MLP) & $[32, 32]$ & Initial Sigma ($\sigma$) & $0.1$ \\
Activation Func. & Tanh & BC Training Epochs & $10$ \\
Total Timesteps & $100,000$ & Demo Trajectories & $100$ \\
Eval. Frequency & $5,000$ & Episode Length & $100$ \\ \hline
\end{tabular}
\end{table}

The evaluation utilizes four distinct baseline categories: (i) \textbf{Random} and \textbf{Static} actions, (ii) a \textbf{Rule-based} reactive controller, (iii) standard \textbf{PPO} trained from scratch, and (iv) the \textbf{CMA-ES oracle}. Crucially, the CMA-ES oracle serves as an empirical offline upper bound; it optimizes trajectories with full foresight of the stochastic input sequence, making it non-deployable for real-time control but ideal for generating high-quality expert demonstrations.

For the \textbf{PPO+BC} variant, the policy was pretrained for 10 epochs on 100 high-performing trajectories generated by the CMA-ES oracle following the procedure in \cite{maus_leveraging_2025} and subsequently fine-tuned via RL. All agents share identical network architectures and seeds to ensure a fair comparison. Evaluation was performed over $20$ independent test seeds (0--19) to capture robustness against stochastic input variations, strictly separated from training (1000+) and demonstration seeds (3000+).

For PPO, we report both the performance of the best intermediate checkpoint (highest evaluation reward) and the final model after 100,000 steps to account for potential training instability. 

The performance distribution across all test seeds is summarized in Fig. \ref{Fig:Results}. Standard PPO outperforms the random baseline but exhibits significant variance and unstable convergence. In several seeds, the final PPO policy even fails to surpass the simple rule-based controller or degrades compared to its own best intermediate checkpoint, highlighting the difficulty of exploring the nonlinear reward landscape from scratch.

In contrast, PPO+BC achieves significantly higher mean returns and demonstrates superior stability. By internalizing the regulation patterns from the CMA-ES demonstrations, the agent avoids early sub-optimal local minima. While the CMA-ES oracle consistently sets the performance peak due to its foresight advantage, PPO+BC approximates this upper bound far more effectively than standard PPO. These results indicate that evolutionary warm-starting not only accelerates learning but also provides a necessary stabilizing effect for industrial continuous control tasks.

\section{Discussion and Conclusion}
This study examined whether evolution strategies can be used to generate high-quality demonstration data for continuous industrial control tasks and whether such demonstrations improve the effectiveness of reinforcement learning. By adapting an existing industrial sorting benchmark \cite{maus_leveraging_2025} to a continuous-control formulation and applying the CMA-ES to optimize episode-long action sequences, we created a controlled setting to evaluate hybrid evolutionary RL approaches.

Across all experiments, the results show a consistent pattern. PPO trained from scratch surpasses naive baselines but exhibits high variance and is occasionally outperformed by simple reactive heuristics. Incorporating CMA-ES-generated demonstrations via behavioral cloning substantially improves both mean performance and robustness. While the per-seed CMA-ES optimization serves as a necessary upper performance baseline, its reliance on fixed-seed foresight prevents deployment in real-time settings. 

Importantly, the CMA-ES oracle is not a competing control solution but a diagnostic and planning tool. 
Its optimization is seed-specific, offline, and scales poorly with episode length and action dimensionality. 
In contrast, PPO+BC incorporates the oracle behavior into a reactive policy that operates without reliance on
foresight and generalizes across unseen input sequences. The objective is therefore not to replace RL, but to stabilize and accelerate it using offline evolutionary planning.

Our results confirm that evolutionary planning on environment simulations inspired by digital twins can successfully encode robust control policies for RL agents. These findings extend earlier work on demonstration-augmented RL in discrete domains \cite{maus_sortingenv_2025, maus_leveraging_2025} to continuous regulation under non-stationary inputs.

While the results are promising, several limitations merit attention. Our benchmark abstracts from physical effects and noise present in real plants. The reward function is tailored to this task and may be adapted when changing the setup, and algorithmic parameters for both PPO and the CMA-ES may be optimized further for higher performance. In particular, the computational cost of the CMA-ES scales with both population size and episode length. These simplifications were deliberate, allowing extensive evaluations across seeds and isolating the impact of demonstration quality without prohibitive computational cost.

Future work should investigate whether the observed benefits transfer to richer environments with longer horizons, additional control dimensions, and more realistic process dynamics. A systematic comparison of computational footprints between RL agents and evolutionary optimizers would also be valuable, especially for use cases with real-time constraints. Extending the benchmark with more detailed physics or multi-objective reward structures could further enhance its practical relevance.

In summary, this work provides a clear proof of concept: evolution strategies can generate high-quality demonstrations that meaningfully accelerate and stabilize RL training in continuous industrial regulation tasks. The benchmark presented here offers a compact foundation for controlled methodological studies, and the insights obtained may support future applied research toward more robust and efficient RL pipelines in industrial settings.

\section*{Acknowledgment}
This research received external funding from the German Federal Ministry for Economic Affairs and Climate Action through the grant “EnSort”. An AI-based language model was used for limited assistance in improving wording, clarity, and code readability. All scientific content, methodology, experimental design, and conclusions were developed by the authors.


\begin{footnotesize}
\bibliographystyle{unsrt}
\bibliography{Bibliography}
\end{footnotesize}


\end{document}